%% file: iclr2027_conference.tex
\documentclass{article} 
\usepackage{iclr2027_conference,times}

\input{math_commands.tex}

\usepackage{hyperref}
\usepackage{url}
\usepackage{algorithm}
\usepackage{algorithmic}
\usepackage{comment}
\usepackage{amssymb}
\usepackage{amsmath}
\usepackage{makecell}
\usepackage{bm}
\usepackage{booktabs}
\usepackage{graphicx}
\usepackage{newfloat}
\usepackage{listings}
\usepackage{multirow}

\title{
FuncRoom-Agent: Sequential Feed-Forward 3D\\
Functional Indoor Scene Generation
}
\iclrfinalcopy
\author{
\small
Hao Feng, Zhi Zuo, MingJian Liang, Jingyu Hu, Xiaowei Hu,\\
Liupengfei Wu, Dian Zhang, Guoxin Fang, Zhengzhe Liu\thanks{Corresponding author.}
}
\begin{document}

\maketitle
\lhead{}

\begin{abstract}
We introduce \textbf{Function-Room Generation}, a new indoor 3D scene generation setting that creates rooms supporting explicit functional goals rather than merely visually plausible layouts. Existing agentic and executable methods improve controllability, but often depend on costly test-time generate--evaluate--revise loops, making functional room generation slow and computationally expensive. We address this challenge with three technical contributions. First, we design a \textbf{recursive domain-specific language} to effectively organize the hierarchical object compositions required by functional rooms, from room structure and major furniture to dense support-surface and nested small objects. It represents rooms as staged executable programs with explicit geometric and functional relations. Second, we propose a \textbf{sequential feed-forward scene construction} framework that distills recursive construction traces into a scene construction expert. At inference time, the expert writes executable DSL code stage by stage, and a deterministic executor directly instantiates each stage without teacher agents, online critics, or iterative repair. Third, we introduce \textbf{ScenePRM}, an execution-grounded process reward framework that improves the expert through reinforcement learning with functional, geometric, relational, and future-constructability feedback. We further establish a function-oriented benchmark and show state-of-the-art performance on both general indoor scene generation and function-room generation, achieving stronger functional completeness, relation correctness, geometric executability, and generation efficiency.
\end{abstract}

\section{Introduction}

Indoor 3D scene generation is important for embodied AI, robotics, virtual reality, simulation, and interactive content creation. A high-quality indoor scene is not merely a collection of visually plausible objects; it must organize many objects into a coherent, usable, and geometrically valid environment. This is challenging because indoor rooms involve dense object compositions, complex spatial relations, hierarchical support structures, wall- and ceiling-mounted objects, and numerous small functional objects. Major furniture defines activity zones, while local objects and support-surface arrangements determine whether the room can actually support a target activity.

Existing indoor scene generation methods broadly follow three paradigms.
Data-driven layout synthesis methods generate plausible object layouts
from learned spatial statistics\citep{paschalidou2021atiss,
tang2024diffuscene}, but mainly target generic rooms and coarse furniture
arrangements, without explicitly modeling the activity zones, functional
objects, and dense small-object compositions required by function-oriented
spaces. Language-guided methods use general-purpose LLMs or VLMs to infer
scene plans and spatial constraints\citep{yang2024holodeck,
sun2025layoutvlm,yang2025optiscene}, but these models are not specialized
for 3D construction and may produce inconsistent scales, invalid support
or wall-relative relations, collisions, inaccessible objects, or
non-executable outputs. More recent agentic and executable systems improve
controllability and validity through code, tools, critics, and execution
feedback\citep{yang2025sceneweaver,scenesmith2026,xia2026sage,
wang2026scenecode,yang2026codeasroom,ran2025direct}, yet their quality often depends on
costly test-time generation, evaluation, repair, or optimization loops,
resulting in slow and computationally expensive inference. These limitations motivate a
function-oriented representation, a specialized 3D scene construction
expert, and an efficient generation paradigm without test-time iterative
repair.

In this paper, we study \textbf{Function-Room Generation}: given a natural-language functional intent, the goal is to generate
an executable indoor 3D room that supports the specified activities.
Beyond visual plausibility, a valid room should contain the required
functional zones and objects and organize them into a usable spatial
layout. We introduce a function-oriented benchmark that evaluates
function-zone coverage, functional-object coverage and precision, and
generation efficiency.

To address this task, we propose a \textbf{sequential feed-forward recursive construction} framework with three key components. First, we introduce a \textbf{recursive domain-specific language (DSL)} tailored to the hierarchical composition of functional rooms. The DSL represents a room as a staged executable program, organizing construction from room structure and activity-defining furniture to wall- and ceiling-mounted objects, support-surface arrangements, and recursively nested local details. It explicitly encodes functional hierarchies, support relations, wall-relative placements, geometric constraints, and asset requests, enabling complex functional object compositions to be represented in a structured form. Second, we develop \textbf{sequential feed-forward scene construction} through visual-context supervised fine-tuning. A general-purpose multimodal code agent first generates recursive DSL construction traces by writing, executing, and rendering DSL segments stage by stage. The resulting scenes and traces are manually reviewed, and then distilled into a compact \textbf{scene construction expert} that learns to progressively translate functional intents into executable DSL segments conditioned on the current partial scene and visual context. At inference time, each generated executable DSL segment is directly executed by a deterministic executor, enabling efficient room construction without test-time generate--evaluate--revise loops. Third, we introduce \textbf{ScenePRM}, an execution-grounded process reward framework that further optimizes the expert through reinforcement learning. ScenePRM evaluates whether intermediate construction states make valid progress toward a complete functional room in terms of functional completeness, relation correctness, support validity, wall consistency, geometric validity, and future constructability. The trained expert therefore performs efficient sequential feed-forward generation without invoking the teacher agent, ScenePRM, or iterative repair at test time.

We evaluate our method on both general indoor scene generation and the proposed Function-Room benchmark, demonstrating state-of-the-art performance in functional completeness, relation correctness, support validity, and geometric executability. Moreover, the sequential feed-forward design reduces generation time to 25 minutes, outperforming the closest baseline by 24.2\% and avoiding the hours-long test-time refinement required by several agentic systems.

Our contributions are summarized as follows:
\begin{itemize}
\item We formulate \textbf{Function-Room Generation} and establish a function-oriented benchmark that evaluates whether generated 3D indoor scenes satisfy explicit functional goals.
\item We design a \textbf{recursive DSL} that represents functional rooms as staged executable programs with explicit functional hierarchies, support relations, wall-relative placements, and geometric constraints.
\item We propose a \textbf{sequential feed-forward scene construction framework} that distills multimodal construction trajectories through visual-context supervised fine-tuning into a scene construction expert, which generates and executes DSL code stage by stage without test-time iterative repair.
\item We introduce \textbf{ScenePRM}, an execution-grounded process reward framework that further optimizes the scene construction expert through reinforcement learning, achieving state-of-the-art performance on both function-oriented and general indoor scene generation.
\end{itemize}

\section{Related Work}

\subsection{Generic Indoor Layout and LLM/VLM Planning}

Indoor scene generation has been widely studied through layout synthesis, scene graphs, and language-guided planning. Data-driven methods represent indoor scenes as object sets, bounding boxes, or relational graphs, and learn to generate plausible layouts conditioned on room types, floor plans, text, or partial observations\citep{paschalidou2021atiss,tang2024diffuscene,hu2024midiffusion,zhai2023commonscenes,pun2025hsm,qi2026globallocalmontecarlotree,yang2024physcene,wang2021sceneformer,yang2021scenesynthesisuncertaintydrivenattribute,liu2023cliplayout}. These methods capture common spatial statistics and object co-occurrence patterns, but they mainly focus on generic layout realism rather than functional usability. Recent language-guided methods further use general-purpose LLMs or VLMs to infer object lists, spatial constraints, scene graphs, or layout plans from natural-language instructions\citep{feng2023layoutgpt,lin2024instructscene,yang2024holodeck,sun2025layoutvlm,yang2025optiscene,xiao2026thinkblox3dindoorscene}. Such models provide useful commonsense priors and improve text controllability, but they are typically used in a zero-shot, prompt-based, or weakly adapted manner. As a result, they are not specialized for 3D indoor construction and may struggle with scale consistency, support hierarchy, wall-relative placement, dense small-object arrangement, and executable scene structure. In contrast, we train a specialized scene construction expert for function-room generation, where the goal is to produce usable rooms that satisfy explicit functional goals.

\subsection{Agentic and Executable Scene Construction}

Recent agentic systems improve 3D scene generation by decomposing scene construction into planning, generation, evaluation, and refinement stages. Representative methods use LLM/VLM agents, generators, critics, renderers, simulators, and retrieval tools to improve semantic alignment, visual realism, physical plausibility, and simulation readiness\citep{yang2025sceneweaver,scenesmith2026,xia2026sage}. These methods show that multi-agent reasoning and critic-guided refinement can produce rich indoor scenes, especially for embodied AI and simulation. However, their quality often depends on test-time tool invocation and iterative generate--evaluate--revise loops.

Executable scene representations provide a complementary direction. Recent methods represent indoor scenes as code, structured programs, or DSLs to support execution, validation, editing, and interaction\citep{wang2026scenecode,yang2026codeasroom,chen2025roompilot,tang2026spatialgrammar,li2026hdsl}. For example, such systems may generate Blender/Python programs, structured asset requests, indoor DSL specifications, BEV-grid layout programs, or hierarchical scene trees. These representations improve controllability and validity, but they are often used as online workspaces for LLM agents: programs are written, executed, diagnosed, repaired, or optimized during inference. In contrast, our recursive DSL is used as the output language of a learned scene code generator. At test time, the controller writes the next executable DSL segment conditioned on the current partial scene and visual context, and a deterministic executor directly updates the scene without iterative repair for the same state.

\begin{figure*}[t]
\centering
\includegraphics[width=0.99\textwidth]{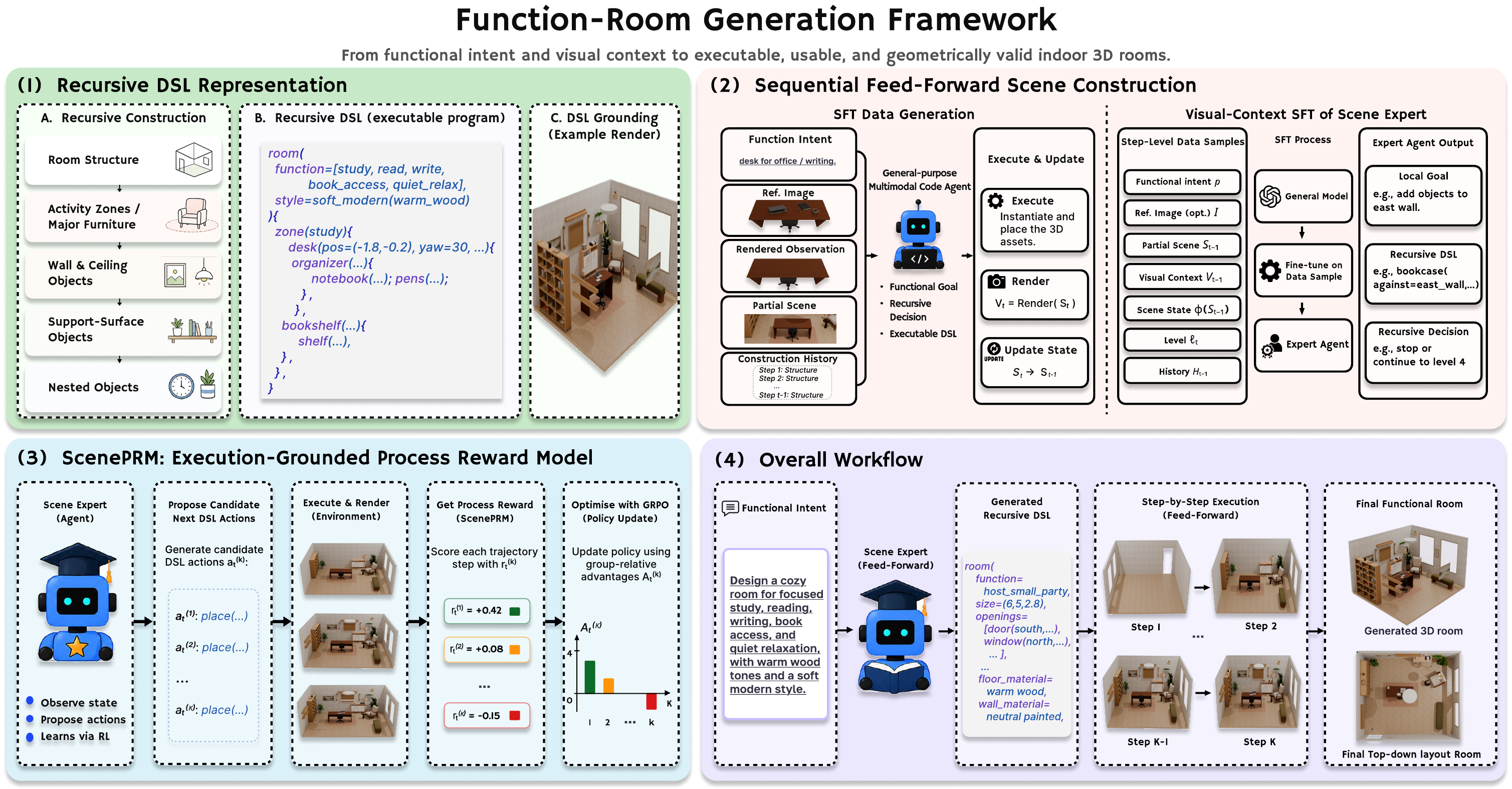} 
\caption{Overview of our Function-Room Generation framework. A recursive DSL enables coarse-to-fine executable scene construction, while visual-context SFT distills verified agent-generated traces into a compact scene expert. ScenePRM further optimizes the expert with execution-grounded GRPO, enabling sequential feed-forward generation of functional 3D rooms at inference time.}
\label{fig:overview}
\end{figure*}

\section{Method}
\label{sec:method}

\subsection{Overview}
\label{sec:method_overview}

Given a natural-language functional intent $p$,
our goal is to generate an executable 3D room that satisfies
functional, relational, and geometric requirements. As shown in
Fig.~\ref{fig:overview}, our framework consists of three key
components. First, a \textbf{recursive DSL} represents the room as a
staged executable program organized into five coarse-to-fine
construction levels. Second, we develop \textbf{sequential
feed-forward scene construction}: a general-purpose multimodal code
agent and an execution harness are used to bootstrap verified recursive
DSL construction traces, which are distilled through visual-context
SFT into a compact scene construction expert. Third,
\textbf{ScenePRM}-guided GRPO further improves the expert by executing,
rendering, and comparing candidate stage rollouts sampled from the same
partial scene. At inference time, the expert generates one executable DSL segment from each partial scene state, which the deterministic executor executes once, without teacher agents, ScenePRM, test-time search, critique, or repair.

\subsection{Function-Room Generation Task Formulation}
\label{sec:problem_formulation}

Given a natural-language functional intent $p$, Function-Room Generation aims to generate an executable indoor
scene $S=\mathcal{G}(p)$ that supports the specified activities.
The prompt describes the intended functions, required objects, and appearance
preferences.

Unlike generic indoor scene generation, the task evaluates whether the generated
scene contains the required functional zones and objects, preserves spatial,
support, and mounting relations, maintains accessibility and circulation, and
satisfies basic geometric constraints such as containment and collision avoidance.

\subsection{Recursive DSL Representation}
\label{sec:recursive_dsl}

Functional rooms contain objects at multiple dependency levels, from
room structure and activity-defining furniture to mounted objects,
support-surface arrangements, and small nested details. A flat object
list cannot explicitly capture these functional and geometric
dependencies. We therefore introduce a recursive DSL that represents
the room as a structured executable program. 

\subsubsection{Recursive Program Tree}
\label{sec:program_tree}

We represent each room as a rooted, ordered program tree
\begin{equation}
\mathcal{T}
=
(\mathcal{V},\mathcal{E}_{\mathrm{tree}}),
\end{equation}
where $\mathcal{V}$ contains function-oriented construction nodes and
$\mathcal{E}_{\mathrm{tree}}$ contains their directed parent--child
relations. Each node is defined as
\begin{equation}
u_i=(g_i,a_i,\rho_i),
\qquad
u_i\in\mathcal{V},
\end{equation}
where $g_i$ is a local functional goal, $a_i$ is the executable DSL
segment that realizes it, and $\rho_i$ is the recursive expansion
decision.

The goal $g_i$ specifies a function group, such as a study area, desk
arrangement, or tabletop reading group. The DSL segment $a_i$
describes the corresponding objects, semantic roles, properties,
placements, constraints, asset requests, and spatial/functional
relations. The expansion decision is
\begin{equation}
\rho_i
\in
\{\texttt{expand},\texttt{leaf}\},
\end{equation}
where \texttt{expand} decomposes the current function group into
finer-grained child groups, and \texttt{leaf} terminates recursion.
For example, a desk arrangement may be expanded into monitor,
reading-tool, and desktop-detail groups. The ordering of child nodes
determines their generation and execution order.

\subsubsection{Recursive Construction Levels}
\label{sec:recursive_levels}

The recursive DSL follows five dependency levels: room structure; activity
zones and major furniture; wall- and ceiling-mounted objects; support-surface
objects; and recursively nested objects.

This ordering ensures that dependent geometry already exists when an object is
generated: mounted objects require valid walls, support-surface objects require
their supporting furniture, and nested objects require their parent objects.
The levels define only the dependency order; the scene construction expert
determines which functional groups are required and how they are recursively
expanded. The complete grammar, operator semantics, geometric constraints,
asset-grounding procedure, and executor implementation are provided in the
supplementary material.

\subsection{Sequential Feed-Forward Scene Construction}
\label{sec:sequential_construction}

We train a compact scene construction expert by first collecting
verified recursive DSL construction traces and then distilling them
through visual-context supervised fine-tuning.

\subsubsection{Recursive Trace Construction}
\label{sec:trace_construction}

We construct recursive scene-building traces using a general-purpose
multimodal code agent coupled with a deterministic execution and
rendering harness. Given the functional intent $p$, current partial scene $S_{t-1}$, rendered observation
$V_{t-1}$, structured scene state $\phi(S_{t-1})$, construction level
$\ell_t$, and preceding history $H_{t-1}$, the agent generates
\begin{equation}
(g_t,a_t,\rho_t)
=
\mathcal{A}
\left(
p,V_{t-1},\phi(S_{t-1}),\ell_t,H_{t-1}
\right),
\end{equation}
where $g_t$ specifies the next local functional goal, $a_t$ is the
executable DSL segment that realizes this goal, and $\rho_t$ determines
whether the current construction node should be recursively expanded.

Each generated DSL segment is immediately parsed, executed, and rendered:
\begin{equation}
(S_t,e_t)
=
\operatorname{Execute}(S_{t-1},a_t),
\qquad
V_t
=
\operatorname{Render}(S_t),
\label{equation:5}
\end{equation}
where $e_t$ denotes the execution status. The updated scene and rendered
observation are then returned to the agent for the next construction
step. Repeating this process produces a state-conditioned trace that
progressively constructs the room from coarse functional structure to
fine local details.

The generated scenes and traces are manually reviewed for functional
completeness, relation correctness, geometric validity, and execution
success; invalid samples are corrected or discarded. The remaining
verified traces are used as supervision for training the scene construction expert.

\subsubsection{Visual-Context Supervised Fine-Tuning}
\label{sec:visual_context_sft}

We decompose each verified construction trace into step-level training
samples:
\begin{equation}
\mathcal{D}_{\mathrm{SFT}}
=
\left\{
(x_t,y_t^{*})
\right\},
\qquad
y_t^{*}
=
(g_t^{*},a_t^{*},\rho_t^{*}),
\end{equation}
where $x_t$ contains the functional intent,
current rendered observation, structured partial-scene state,
construction level, and preceding construction history. The target
$y_t^{*}$ contains the next local functional goal, its executable DSL
segment, and the recursive expansion decision.

We use these state-conditioned code-generation samples to distill the
general-purpose multimodal code agent into a compact scene construction
expert. The supervised objective is
\begin{equation}
\mathcal{L}_{\mathrm{SFT}}
=
-
\sum_{(x,y^{*})\in\mathcal{D}_{\mathrm{SFT}}}
\sum_{j=1}^{|y^{*}|}
\log
\pi_{\theta}
\left(
y_j^{*}\mid x,y_{<j}^{*}
\right).
\end{equation}
This training teaches the expert to generate the next executable DSL
segment directly from the current executed scene state and visual
context, enabling sequential feed-forward construction at inference
time.

\subsection{ScenePRM-Guided Reinforcement Learning}
\label{sec:sceneprm}

Supervised fine-tuning enables the compact scene construction expert
to imitate verified traces, but token-level likelihood training does
not ensure that each generated DSL segment advances the target room
function. A segment may be executable and geometrically plausible
while still omitting required functional objects, producing incomplete
activity zones, or introducing relations that hinder later
construction. We therefore introduce \textbf{ScenePRM}, an
execution-grounded process reward framework that further optimizes the
SFT expert using group-relative policy optimization. ScenePRM evaluates
whether each construction stage makes measurable progress toward a
complete, usable, and geometrically valid functional room.

Given the accepted partial scene before construction level $\ell$, we
sample $K$ candidate stage rollouts from the current policy. The
$k$-th rollout is defined as
\begin{equation}
\tau_{\ell}^{(k)}
=
\left(y_t^{(k)}\right)_{t\in\mathcal{I}_{\ell}},
\qquad
k=1,\ldots,K,
\end{equation}
where
\begin{equation}
y_t^{(k)}
=
\left(
g_t^{(k)},a_t^{(k)},\rho_t^{(k)}
\right)
\sim
\pi_{\theta}
\left(
\cdot\mid x_t^{(k)}
\right).
\end{equation}
Here, $\mathcal{I}_{\ell}$ contains the recursive construction steps
belonging to level $\ell$. All candidates start from the same partial
scene and independently generate the content required at that level.

Each candidate rollout is executed step by step following
Eq.~\ref{equation:5}, yielding a terminal scene $S_{\ell}^{(k)}$, its
rendered observation $V_{\ell}^{(k)}$, and a stage-level
execution status $e_{\ell}^{(k)}$.

ScenePRM uses a VLM-based evaluator to score each successfully
executed candidate along four dimensions:
\[
\mathcal{M}
=
\{
\mathrm{real},
\mathrm{func},
\mathrm{layout},
\mathrm{prompt}
\},
\]
corresponding to visual realism, functionality, layout quality, and
prompt following. Its average quality score is
\begin{equation}
\bar{q}_{\ell}^{(k)}
=
\frac{1}{|\mathcal{M}|}
\sum_{m\in\mathcal{M}}
E_m
\left(
p,
S_{\ell}^{(k)},
V_{\ell}^{(k)},
\ell
\right).
\end{equation}
The functionality and layout evaluators jointly assess functional
completeness, relation correctness, support validity, wall
consistency, geometric validity, and future constructability.

The stage-level process reward is defined as
\begin{equation}
r_{\ell}^{(k)}
=
\begin{cases}
-1,
&
e_{\ell}^{(k)}
\neq
\texttt{success},
\\[3pt]
\bar{q}_{\ell}^{(k)},
&
e_{\ell}^{(k)}
=
\texttt{success}.
\end{cases}
\end{equation}
Thus, execution-failed candidates are directly penalized, while
successful candidates are scored according to the quality of the
resulting partial scene.

Because all $K$ candidates are sampled from the same construction
state, we compute their group-relative advantages as
\begin{equation}
A_{\ell}^{(k)}
=
\frac{
r_{\ell}^{(k)}-\mu_{\ell}
}{
\sigma_{\ell}+\epsilon
},
\end{equation}
where $\mu_{\ell}$ and $\sigma_{\ell}$ are the mean and standard
deviation of the candidate rewards at level $\ell$. We optimize the
scene construction expert using the standard clipped GRPO objective
with KL regularization against the frozen SFT reference policy.
During recursive training, the highest-reward candidate is used to
initialize the next construction level. Candidate sampling, ScenePRM
evaluation, and candidate selection are used only during training and
are removed at inference time.

\subsection{Sequential Feed-Forward Inference}
\label{sec:feedforward_inference}

At inference time, we retain only the trained scene construction expert,
deterministic executor, and renderer. Given the current construction state
$x_t$, the expert directly predicts the next output
$(g_t,a_t,\rho_t)$, consisting of a local functional goal, an executable DSL
segment, and a recursive expansion decision. Following
Eq.~\ref{equation:5}, the DSL segment is executed and rendered once, and the
updated scene state and visual observation condition the next prediction. The
expansion decision determines whether construction continues within the
current functional group, recursively expands it, or proceeds to the next
level.

We call this process \emph{sequential feed-forward inference}: it is sequential
because each prediction depends on the previously executed partial scene, and
feed-forward because each state undergoes one model generation and one
deterministic execution. The teacher agent, ScenePRM, and candidate sampling
are used only during training.

\section{Experiments}
\label{sec:experiments}

\begin{figure*}[t]
    \centering
    \makebox[\textwidth][c]{%
        \includegraphics[width=1\textwidth]{Figures/qualitative_comparisonV8.pdf}%
    }
    \caption{
        Qualitative comparison on the Function-Room benchmark.
        Each column represents a different scene-generation method,
        and each row shows results generated from the same functional prompt.
    }
    \label{fig:function_qualitative}
\end{figure*}

\subsection{Experimental Setup}
\label{sec:experimental_setup}

\paragraph{Base model and training.}
We use Qwen3.6-35B-A3B as the base model. The model is first trained with supervised fine-tuning (SFT) using data generated by Codex with GPT-5.5, and is then further optimized with ScenePRM-guided agentic reinforcement learning. Complete training and inference details, including SFT and reinforcement-learning hyperparameters and ScenePRM evaluator configurations, are provided in the appendix.

\paragraph{Benchmarks.}
We evaluate our method on two benchmarks. On SceneEval\citep{tam2026sceneeval}, we follow SceneCode\citep{wang2026scenecode} by using the same 30 room-level prompts and report nine metrics covering textual consistency, physical plausibility, and spatial usability. We also construct a Function-Room benchmark with 50 prompts excluded from training, each specifying multiple functional zones and the furniture and equipment needed to support them. Besides function-specific metrics, we use selected annotation-free SceneEval metrics to assess the general plausibility and visual quality of the generated rooms. Further details on benchmark construction and evaluation protocols, together with additional reference-based metrics such as FID, are provided in the appendix for a more comprehensive evaluation.

\paragraph{Baselines and evaluation protocol.}
On SceneEval, we compare against HSM, LayoutVLM, SceneWeaver, SceneSmith, SceneCode, and Code as Room. On the Function-Room benchmark, we compare against SceneWeaver, SceneSmith, SceneCode, Code as Room and SAGE. All methods use the same prompts and evaluation criteria. For each prompt, an LLM identifies the target functional zones, required object slots, valid alternatives, and maximum valid counts, and a VLM evaluates each generated scene against these prompt-specific requirements.  Detailed evaluation configurations are provided in the appendix.

\begin{table*}[ht]
    \centering
    \caption{Room-level quantitative results on SceneEval.}
    \label{tab:sceneeval}

    \small
    \setlength{\tabcolsep}{2.0pt}
    \renewcommand{\arraystretch}{1.05}

    \resizebox{\textwidth}{!}{
    \begin{tabular}{lcccccccccc}
    \toprule
    Method
    & \#Obj
    & CNT$\uparrow$
    & ATR$\uparrow$
    & OOR$\uparrow$
    & OAR$\uparrow$
    & SUP$\uparrow$
    & ACC$\uparrow$
    & NAV$\uparrow$
    & COL$\downarrow$
    & OOB$\downarrow$
    \\
    \midrule

    HSM
    & $21.2_{\pm2.4}$
    & $53.3_{\pm9.3}$
    & $58.1_{\pm14.7}$
    & $22.3_{\pm12.3}$
    & $50.0_{\pm10.4}$
    & $70.9_{\pm7.2}$
    & $85.8_{\pm4.8}$
    & $99.1_{\pm1.0}$
    & $18.4_{\pm6.8}$
    & $7.4_{\pm3.6}$
    \\

    LayoutVLM
    & $12.6_{\pm1.7}$
    & $64.2_{\pm8.0}$
    & $31.1_{\pm13.7}$
    & $27.2_{\pm12.1}$
    & $53.1_{\pm14.8}$
    & $25.7_{\pm7.7}$
    & $93.3_{\pm5.2}$
    & $\bm{100.0_{\pm0.0}}$
    & $21.1_{\pm10.0}$
    & $5.9_{\pm3.5}$
    \\

    SceneWeaver
    & $18.4_{\pm2.3}$
    & $58.9_{\pm11.2}$
    & $49.3_{\pm12.9}$
    & $27.9_{\pm13.6}$
    & $51.3_{\pm10.1}$
    & $46.4_{\pm6.2}$
    & $86.4_{\pm8.4}$
    & $93.3_{\pm5.7}$
    & $25.9_{\pm14.7}$
    & $1.6_{\pm0.3}$
    \\

    SceneSmith
    & $62.5_{\pm25.4}$
    & $78.8_{\pm6.7}$
    & $71.6_{\pm9.4}$
    & $37.2_{\pm13.3}$
    & $72.4_{\pm10.3}$
    & $66.7_{\pm4.0}$
    & $52.2_{\pm8.7}$
    & $97.8_{\pm2.1}$
    & $18.1_{\pm4.8}$
    & $0.6_{\pm0.7}$
    \\

    SceneCode
    & $32.2_{\pm10.3}$
    & $79.4_{\pm6.0}$
    & $74.0_{\pm11.7}$
    & $32.6_{\pm12.6}$
    & $61.1_{\pm12.6}$
    & $44.3_{\pm12.8}$
    & $74.2_{\pm9.1}$
    & $\bm{100.0_{\pm0.0}}$
    & $11.3_{\pm4.3}$
    & $0.4_{\pm0.9}$
    \\

    Code as Room
    & $23.2_{\pm5.3}$
    & $57.2_{\pm16.1}$
    & $59.8_{\pm15.5}$
    & $16.7_{\pm13.2}$
    & $46.9_{\pm10.3}$
    & $27.3_{\pm13.7}$
    & $56.1_{\pm6.7}$
    & $99.9_{\pm0.1}$
    & $28.9_{\pm4.3}$
    & $0.7_{\pm0.5}$
    \\

    \midrule

    Ours
    & $\bm{64.3_{\pm17.2}}$
    & $\bm{82.4_{\pm9.5}}$
    & $\bm{76.8_{\pm11.3}}$
    & $\bm{38.1_{\pm15.1}}$
    & $\bm{74.0_{\pm13.9}}$
    & $\bm{72.8_{\pm8.3}}$
    & $\bm{93.5_{\pm10.6}}$
    & $\bm{100.0_{\pm0.0}}$
    & $\bm{10.7_{\pm7.8}}$
    & $\bm{0.3_{\pm0.2}}$
    \\

    \bottomrule
    \end{tabular}
    }
\end{table*}

\begin{table*}
    \centering
    \caption{Quantitative results on the Function-Room benchmark. Real., Aesth., and Func. denote the user-study scores for realism, aesthetics, and functionality, respectively.  Generation time is recorded in seconds and converted into hours and minutes for readability.}
    \label{tab:function_room}
    \setlength{\tabcolsep}{2pt}
     \resizebox{\textwidth}{!}{
    \begin{tabular}{ccccccccccccc}
        \toprule
        Method
        & FZC$\uparrow$
        & FOC$\uparrow$
        & FOP$\uparrow$
        & ACC$\uparrow$
        & SUP$\uparrow$
        & NAV$\uparrow$& OOB$\downarrow$& COL$\downarrow$
        & Real.$\uparrow$
        & Aesth.$\uparrow$
        & Func.$\uparrow$
        & Gen. Time$\downarrow$ \\
        \midrule
        SAGE& 43.6& 48.7& 56.5& 53.0& 52.4& 98.1& 13.9& 16.8& 5.3& 5.5& 4.7&2h16m\\
        SceneWeaver
        & 45.0
        & 73.4
        & 61.4
        & 35.4
        & 51.6
        & 97.6
        & 12.5
        & 38.1
        & 4.3
        & 4.7
        & 5.6
        & 1h14m \\
        
        SceneSmith
        & 68.8
        & 69.4
        & 44.7
        & 51.6
        & 51.2
        & 93.6
        & 23.1
        & 18.9
        & 6.2
        & 7.5
        & 6.8
        & 4h15m \\
        
        SceneCode
        & 47.5
        & 50.3
        & 58.3
        & 51.4
        & 51.7
        & \textbf{100.0}
        & 33.7
        & {15.6}& 5.6
        & 6.4
        & 5.1
        & 6h16m \\
        
        Code as Room
        & 72.9
        & 82.5
        & 63.2
        & 33.5
        & 59.8
        & 99.9
        & 15.7
        & 37.1
        & 5.5
        & 5.8
        & 6.2
        & 33m \\
        \midrule
        
        Ours
        & \textbf{82.5}& \textbf{86.3}& \textbf{65.7}& \textbf{69.1}& \textbf{68.8}& \textbf{100.0}
        & \textbf{11.2}& \textbf{14.5}& \textbf{7.4}& \textbf{8.3}& \textbf{7.9}& \textbf{25m} \\
        \bottomrule
        \end{tabular}
        }
\end{table*}

\begin{table*}[h!]
    \centering

    \begin{minipage}[t]{0.54\textwidth}
        \centering
        \caption{Comparison of one-shot, staged, and recursive agentic construction strategies on the Function-Room benchmark.}

        \setlength{\tabcolsep}{1pt}

        \resizebox{\linewidth}{!}{
        \begin{tabular}{ccccccccc}
        \toprule
        Method
        & FZC$\uparrow$
        & FOC$\uparrow$
        & FOP$\uparrow$
        & ACC$\uparrow$
        & SUP$\uparrow$
        & NAV$\uparrow$
        & OOB$\downarrow$
        & COL$\downarrow$ \\
        \midrule

        One-shot
        & 37.6
        & 34.3
        & 45.6
        & 28.5
        & 39.2
        & 76.7
        & 45.1
        & 48.2 \\

        Staged
        & 42.3
        & 40.9
        & 38.6
        & 30.3
        & 44.1
        & 89.4
        & 39.2
        & 40.5 \\

        Recursive
        & 49.5
        & 58.8
        & 40.4
        & 34.6
        & 46.2
        & 95.4
        & 36.7
        & 37.9 \\

        \bottomrule
        \end{tabular}
        \label{tab:DSL_ablation}
        }

    \end{minipage}
    \hfill
    \begin{minipage}[t]{0.43\textwidth}
        \centering
        \caption{Ablation of training components on the Function-Room benchmark.}

        \setlength{\tabcolsep}{2pt}

        \resizebox{\linewidth}{!}{
        \begin{tabular}{ccccccc}
        \toprule
        Method
        & SFT
        & \makecell{Scene-level\\RL}
        & ScenePRM
        & FZC$\uparrow$
        & FOC$\uparrow$
        & FOP$\uparrow$ \\
        \midrule

        A
        & 
        & 
        & 
        & 50.8
        & 55.2
        & 34.7 \\

        B
        & \checkmark
        & 
        & 
        & 58.9
        & 69.4
        & 44.2 \\

        C
        & \checkmark
        & \checkmark
        & 
        & 76.4
        & 73.0
        & 51.9 \\

        D
        & \checkmark
        & 
        & \checkmark
        & \textbf{82.5}
        & \textbf{86.3}
        & \textbf{65.7} \\

        \bottomrule
        \end{tabular}
        \label{tab:PRM_ablation}
        }

    \end{minipage}

\end{table*}

\paragraph{User study.}
We also conduct a blinded user study on the Function-Room benchmark to assess perceptual quality and practical usability. The study includes 15 participants: five professional interior designers and ten non-experts. Participants view each functional prompt and its generated scenes without method names, model identities, or generation details, and rate them in terms of realism, aesthetics, and functionality. This protocol provides an evaluator-independent assessment separate from both the training-time ScenePRM evaluator and the automatic benchmark evaluator. The aggregated Real., Aesth., and Func. scores are reported in Table~\ref{tab:function_room}, with full procedures and instructions provided in the appendix.

\subsection{Function Room Evaluation Metrics}
\label{sec:function_metrics}

For each benchmark prompt, we construct a fixed set of functional zones,
functional-object slots, valid alternatives, and relevant object
categories. These annotations are initially proposed by an LLM and then
manually reviewed and corrected. The resulting annotations are frozen
and shared across all evaluated methods. \textbf{Function Zone Coverage (FZC)} measures the coverage of the
annotated functional zones. \textbf{Functional Object Coverage (FOC)} measures the coverage of the
annotated functional-object slots, allowing predefined functionally
equivalent alternatives. \textbf{Functional Object Precision (FOP)} measures the proportion of
generated objects that are valid and relevant under the fixed
prompt-specific annotation, while penalizing irrelevant or excessive
objects. \textbf{Generation Time (GT)} measures the wall-clock time required to
complete scene generation under a unified configuration.
Detailed definitions and the annotation protocol are provided in the
appendix.


\subsection{Quantitative Results on SceneEval}
\label{sec:sceneeval_results}

Table~\ref{tab:sceneeval} reports the room-level results on SceneEval. Our
method achieves the best CNT, ATR, OOR, OAR, SUP, and ACC scores, ties for the
best NAV score, and obtains the lowest COL and OOB rates. Notably, it generates
an average of 64.3 objects per room, the highest among the evaluated methods,
while maintaining the strongest accessibility, collision avoidance, and
boundary compliance. This result indicates that the proposed recursive
construction process supports dense functional compositions without
sacrificing geometric usability.

\subsection{Quantitative Results on Function Rooms}
\label{sec:function_results}

As shown in Table~\ref{tab:function_room}, our method achieves the highest FZC,
FOC, and FOP scores of 82.5, 86.3, and 65.7, respectively, demonstrating
improved functional-zone completeness, required-object coverage, and content
precision. It also achieves the best ACC and SUP scores, ties for the best NAV
score, and obtains the lowest OOB and COL rates. In the blinded user study, our
method receives the highest realism, aesthetics, and functionality scores of
7.4, 8.3, and 7.9. The reported 25-minute runtime per scene includes asset generation and retrieval;
the sequential feed-forward construction itself is substantially faster and
constitutes only a small fraction of the total runtime. Detailed runtime
breakdowns and comparisons are provided in the appendix.

\subsection{Ablation Study}
\label{sec:ablation}

\paragraph{Agentic construction strategy ablation.}
Table~\ref{tab:DSL_ablation} evaluates the general-purpose
multimodal code agent on the same 50 prompts without
task-specific SFT or reinforcement learning, following the
construction procedure used for data generation. Staged
construction improves functional coverage over one-shot
generation, while recursive construction further improves
FZC, FOC, and all reported spatial metrics, showing the benefit
of recursive organization beyond coarse-to-fine staging.
These results validate the design of our recursive DSL: explicitly
decomposing functional rooms into hierarchical construction groups
provides a more effective structure for organizing complex scene
generation. In particular, the consistent gains in both functional
coverage and spatial metrics suggest that the recursive representation
helps coordinate semantic completeness with geometric validity.
One-shot obtains a higher FOP largely because it generates substantially
fewer objects, favoring precision at the cost of much lower coverage.
These pre-training results are not directly comparable to those of the
trained scene expert. Qualitative comparisons are provided in the appendix.

\paragraph{Training component ablation.}
Table~\ref{tab:PRM_ablation} shows that SFT mainly improves functional-object
coverage. Starting from the SFT model, our standard RL baseline adopts GRPO,
which samples complete construction traces and optimizes the policy using only
final-scene rewards. This baseline mainly improves functional-zone coverage,
whereas ScenePRM further raises FZC, FOC, and FOP by 6.1, 13.3, and 13.8
points, respectively. These results indicate that ScenePRM-guided optimization
is better suited to long-horizon agentic scene construction than GRPO with
outcome-only rewards. By providing process-level feedback on intermediate
construction states, ScenePRM enables more informative credit assignment for
the sequential decisions that determine the final scene quality.

\subsection{Qualitative Results}
\label{sec:qualitative_results}

Qualitative Results. Figure~\ref{fig:function_qualitative} presents representative qualitative comparisons. Typical baseline errors include a floor lamp blocking the line of sight required for vision testing, a sofa facing the wall rather than the bowling activity area, and misalignment between the MRI scanner and patient bench. Our method generally produces more complete functional regions and more appropriate object orientations and spatial relations. More examples are provided in the supplementary material. 

\section{Conclusion}

We introduced \textbf{Function-Room Generation}, a new setting that shifts indoor 3D scene generation from generic visual plausibility toward explicit functional usability. Our framework combines a recursive DSL for hierarchical functional object compositions, sequential feed-forward scene construction, and ScenePRM-guided reinforcement learning with execution-grounded process feedback. By distilling multimodal construction traces and training a specialized scene construction expert, our method moves costly planning and refinement from inference time to training time, enabling direct and efficient generation without iterative repair. Experiments on SceneEval and the proposed Function-Room benchmark demonstrate improved functional completeness, spatial and support relations, perceptual quality, and generation efficiency, advancing indoor scene generation toward environments structured around how they are intended to be used.

\subsection*{AI use statement}

In this work, we used generative AI tools to assist with dataset construction and parts of the automated evaluation. Generative AI was used to support the construction of data for the generative task and, in some evaluation procedures, as a model-based evaluator to automatically assess or score generated outputs. All AI-assisted data were checked against predefined data specifications and quality criteria. For evaluations involving generative AI, we used standardized prompts, evaluation criteria, and evaluation procedures, and manually reviewed a subset of the results to assess the consistency and reliability of the evaluation.
We also used generative AI tools to assist with translation, language polishing, and improving the readability of parts of the manuscript. Generative AI was not used to independently formulate the core conclusions of this work. All final datasets, experimental results, analyses, and manuscript content were reviewed and verified by the authors. We take full responsibility for the final content of this work, including any data, evaluation results, text, or other artifacts produced with the assistance of generative AI.

\subsection*{Ethics statement}

Our work focuses on function-aware 3D indoor scene generation using publicly
available datasets. Generated scenes may still contain geometric, functional, or safety-related
errors and should therefore not be directly deployed in safety-critical
environments without expert validation. We view the proposed system as a
research tool for studying function-aware scene generation rather than a
replacement for professional spatial or safety assessment.

\subsection*{Reproducibility statement}

We provide the model design and training procedure in the main paper, with
implementation details, training configurations, and evaluation protocols
documented in the appendix. The supplementary materials include the complete
DSL specification and grammar, detailed Function-Room benchmark cases, and
additional qualitative visualizations. Code and evaluation scripts will be
released upon publication.

\bibliography{iclr2027_conference}
\bibliographystyle{iclr2027_conference}

\appendix
\section*{Appendix}

\section{Additional Related Work}
\label{sec:appendix_related_work}

\subsection{Process Supervision for Specialized Scene Controllers}

Training long-horizon agents requires assigning credit to intermediate decisions. Outcome reward models only evaluate final success, while process reward models provide step-level supervision by estimating whether an intermediate decision makes meaningful progress toward the goal\citep{xi2025agentprm,choudhury2025agentprm}. Recent agentic RL methods study credit assignment over multi-turn trajectories, implicit step rewards, and tree-structured rollouts for training LLM agents\citep{luo2025agentlightning,liu2025istar,ji2025treegrpo}. These methods mainly target general tool-use, web navigation, QA, retrieval, or code agents.

Functional indoor scene generation provides a more structured source of process supervision: each partial scene can be parsed, executed, rendered, and evaluated with geometric, relational, visual, and functional checks. We introduce ScenePRM to use such execution-grounded feedback for training the scene construction expert, while keeping inference as sequential feed-forward DSL code generation rather than online critic-and-repair.

\section{Training and Implementation Details}
\label{sec:supp_training_details}

All experiments are conducted on two NVIDIA RTX PRO 6000 GPUs. Our training
set is constructed from 300 room prompts and contains 17,386 recursive traces
and 7,094 grouped traces. Each grouped trace aggregates multiple recursive nodes that belong to the same functional zone, wall-associated structure, or supporting-object hierarchy,
and is stored as a single JSONL entry. The scene construction
expert is initialized from Qwen3.6-35B-A3B and optimized using LoRA with rank
$r=4$, scaling factor $\alpha=8$, and dropout 0.05.

For visual-context supervised fine-tuning, we use a batch size of 2, 8
gradient-accumulation steps, and a maximum sequence length of 8,192, resulting
in an effective batch size of 16. The learning rate is set to $1\times10^{-4}$.

For ScenePRM-guided reinforcement learning, the SFT checkpoint is used for
initialization. At each construction stage, we sample 8 candidate actions and
evaluate them using the deterministic validator and a GPT-5.2-based critic. We
use 16 gradient-accumulation steps, a maximum sequence length of 8,192, and a
learning rate of $8\times10^{-6}$. During rollout generation, the maximum output
length is 2,048 tokens, with temperature 0.8 and top-$p$ sampling with
$p=0.95$. For GRPO, we use group-wise z-score reward normalization, a clipping
range of 0.2, and a KL coefficient of 0.04.

\section{Function Room Evaluation Details}
\label{sec:supp_function_metrics}

\subsection{Benchmark Annotation and Evaluation Protocol}
\label{sec:supp_evaluation}

To ensure that all generated scenes are evaluated against the same
functional requirements, we construct a fixed annotation for each
benchmark prompt before evaluating any method.

Given a prompt $p$, an LLM is used once to propose a set of target
functional zones and functional-object requirements. The resulting
annotation is then manually reviewed, filtered, and corrected to remove
ambiguous, redundant, or functionally irrelevant requirements. After
manual verification, the annotation is frozen and shared across all
evaluated methods and generated scenes. The LLM is therefore not invoked
to regenerate functional requirements during scene evaluation.

For each prompt $p$, the fixed annotation is represented as
\begin{equation}
    \mathcal{A}_p
    =
    \left(
        \mathcal{Z}_p,
        \mathcal{S}_p,
        \mathcal{C}^{p}_{\mathrm{rel}},
        \mathcal{U}_p
    \right),
\end{equation}
where
\begin{equation}
    \mathcal{Z}_p
    =
    \{z_i\}_{i=1}^{N_z}
\end{equation}
is the manually verified set of target functional zones, and
\begin{equation}
    \mathcal{S}_p
    =
    \{s_j\}_{j=1}^{N_s}
\end{equation}
is the set of required functional-object slots. Each slot may contain
multiple manually verified alternatives that provide the same function.
$\mathcal{C}^{p}_{\mathrm{rel}}$ denotes the set of object categories
considered functionally relevant to prompt $p$, and
$\mathcal{U}_p=\{u_c\}$ specifies the maximum valid count for each
relevant category.

During evaluation, a VLM examines each generated scene against this
fixed prompt-specific annotation. It determines whether each annotated
functional zone is present, whether each functional-object slot is
matched, and which generated objects belong to the fixed set of relevant
categories. Thus, all methods are evaluated using identical functional
requirements and acceptance criteria.

\subsection{Function Room Metrics}
\label{sec:supp_function_metrics_definition}

\paragraph{Function Zone Coverage.}
Function Zone Coverage (FZC) measures the proportion of manually verified
functional zones that are present in the generated scene:
\begin{equation}
    \mathrm{FZC}
    =
    \frac{1}{N_z}
    \sum_{i=1}^{N_z}
    \mathbb{I}(z_i),
    \label{eq:supp_fzc}
\end{equation}
where $\mathbb{I}(z_i)\in\{0,1\}$ is the VLM judgment of whether the
annotated zone $z_i$ is present and whether its objects and spatial
organization can support the intended activity. The set of evaluated
zones is fixed for each prompt and is not regenerated for individual
scenes.

\paragraph{Functional Object Coverage.}
Functional Object Coverage (FOC) measures the proportion of manually
verified functional-object slots that are matched:
\begin{equation}
    \mathrm{FOC}
    =
    \frac{1}{N_s}
    \sum_{j=1}^{N_s}
    \mathbb{I}(s_j),
    \label{eq:supp_foc}
\end{equation}
where $\mathbb{I}(s_j)\in\{0,1\}$ indicates whether slot $s_j$ is
matched in the generated scene. A slot may contain multiple predefined
object alternatives that provide the same function, and it is considered
matched when at least one valid alternative is present. Both the slots
and their valid alternatives are fixed after manual verification.

\paragraph{Functional Object Precision.}
Functional Object Precision (FOP) measures the proportion of generated
objects that are valid and functionally relevant under the fixed
prompt-specific annotation. Let $\mathcal{C}^{p}_{\mathrm{rel}}$ denote
the manually verified set of relevant object categories for prompt $p$.
For each category $c\in\mathcal{C}^{p}_{\mathrm{rel}}$, let $n_c$ be
the number of generated instances and $u_c$ be the predefined maximum
valid count. The number of valid and relevant objects is
\begin{equation}
    N_{\mathrm{valid\text{-}rel}}
    =
    \sum_{c\in\mathcal{C}^{p}_{\mathrm{rel}}}
    \min(n_c,u_c).
    \label{eq:supp_valid_relevant}
\end{equation}
FOP is computed as
\begin{equation}
    \mathrm{FOP}
    =
    \frac{N_{\mathrm{valid\text{-}rel}}}
    {N_{\mathrm{eval}}},
    \label{eq:supp_fop}
\end{equation}
where $N_{\mathrm{eval}}$ is the total number of evaluable furniture,
equipment, and functional objects identified in the generated scene.
Architectural elements such as walls, floors, ceilings, doors, and
windows are excluded.

The predefined upper bound $u_c$ prevents excessive duplicate objects
from being counted as additional valid content. Objects outside the fixed
relevant-category set, as well as instances exceeding their corresponding
maximum valid counts, contribute to the denominator but not to
$N_{\mathrm{valid\text{-}rel}}$. FOP therefore penalizes irrelevant and
excessive objects, whereas FOC evaluates whether the required functional
object slots are covered.

\paragraph{Generation Time.}
Generation Time (GT) measures the wall-clock time required to complete
scene generation:
\begin{equation}
    \mathrm{GT}
    =
    t_{\mathrm{completed}}
    -
    t_{\mathrm{submitted}},
    \label{eq:supp_generation_time}
\end{equation}
where $t_{\mathrm{submitted}}$ is the time at which the generation request
is submitted and $t_{\mathrm{completed}}$ is the time at which the
completed scene is returned. Generation time is recorded under the same
generation configuration for all evaluated methods.

\subsection{SceneEval Metrics}
\label{sec:supp_sceneeval_metrics_definition}

We additionally evaluate generated scenes using nine room-level metrics
from SceneEval\citep{tam2026sceneeval}. These metrics assess prompt
consistency, spatial usability, and basic physical plausibility. Higher
values indicate better performance for CNT, ATR, OOR, OAR, SUP, ACC,
and NAV, whereas lower values are preferred for COL and OOB.

\paragraph{Object Count.}
Object Count (CNT) measures the satisfaction rate of object-number
requirements specified by the prompt. Generated objects are first matched
to the corresponding prompt categories, after which their instance counts
are compared with the annotated exact or relative quantity constraints.

\paragraph{Object Attribute.}
Object Attribute (ATR) measures how often generated objects satisfy their
requested visual or semantic properties, such as color, material, and
scale-related descriptions. The required attributes are evaluated from
rendered object views using a VLM.

\paragraph{Object--Object Relationship.}
Object--Object Relationship (OOR) measures the satisfaction rate of
prompt-specified spatial relations between pairs of objects. Relations
such as \emph{in front of}, \emph{beside}, or \emph{on top of} are
verified using the object poses and geometric extents.

\paragraph{Object--Architecture Relationship.}
Object--Architecture Relationship (OAR) measures whether objects are
placed consistently with architectural elements, such as walls, floors,
and ceilings. It evaluates constraints including placement against a
wall, attachment to a wall, or positioning near the center of the room.

\paragraph{Support.}
Support (SUP) measures the proportion of objects with valid physical
support. Depending on the expected support type, an object should be
properly supported by the floor, another object, a wall, or the ceiling.
Support validity is checked using the scene geometry and ray-based
intersection tests.

\paragraph{Accessibility.}
Accessibility (ACC) measures the proportion of objects whose functional
interaction regions remain unobstructed. For example, the front of a
chair or sofa and the usable sides of a bed should retain sufficient free
space for access and interaction.

\paragraph{Navigability.}
Navigability (NAV) evaluates the connectivity of the available floor
space. It is computed as the ratio between the largest connected free-space
component and the total free floor area. A high value indicates that the
scene preserves a largely connected circulation region rather than
fragmenting it into isolated areas.

\paragraph{Collision.}
Collision (COL) measures the proportion of objects involved in at least
one geometric intersection with another object. Pairwise mesh collision
tests are used to identify invalid object overlap. Lower COL indicates
better geometric validity.

\paragraph{Out of Bounds.}
Out of Bounds (OOB) measures the proportion of objects that extend beyond
the valid room footprint. An object is treated as out of bounds when its
geometry is not sufficiently contained above the floor-plan region.
Lower OOB indicates better compliance with the room boundary.

\subsection{User Study Protocol}
\label{sec}

\paragraph{Participants.}
We recruited 15 participants for the user study, including five professional interior designers and ten non-expert participants. The professional designers provided assessments informed by practical interior-design experience, while the non-expert participants reflected the perceptual preferences and usability judgments of general users. Before the evaluation, all participants were informed of the study procedure, the scoring scale, and the definitions of the evaluation criteria.

\paragraph{Evaluation data.}
The user study was conducted on the Function-Room benchmark. We included 50 generated scenes from each evaluated method. Each scene was generated according to a functional prompt describing the activities or requirements that the room should support.

For each scene, participants were shown the corresponding functional prompt together with five rendered views of the generated room: one top-down view and four side views captured from the front, back, left, and right directions. The top-down view provided an overview of the global spatial layout, while the four side views allowed participants to inspect furniture placement, object relationships, accessibility, and potential occlusions from different directions. This multi-view presentation enabled a more comprehensive assessment than evaluation based on a single rendered image.

\paragraph{Blinded evaluation protocol.}
We adopted a blinded evaluation protocol to prevent participants' judgments from being influenced by prior knowledge of the evaluated methods. All generated scenes were associated with anonymous identifiers, and participants were not provided with method names, model identities, implementation details, or any information about how the scenes were produced.

To maintain a manageable evaluation workload, we used a balanced task-assignment strategy rather than requiring every participant to evaluate the complete set of scenes. Each participant was assigned 20 prompt scenes. The assignments were balanced across participants so that the evaluated prompts and generated results were evenly covered. For each assigned prompt, participants evaluated the corresponding generated scenes using the same interface and evaluation instructions. The method identities remained hidden throughout the study.

This evaluation was conducted independently of both the training-time ScenePRM evaluator and the automatic evaluator used for the Function-Room benchmark. Therefore, the user-study results provide an independent human assessment of the perceptual quality and practical usability of the generated scenes.

\paragraph{Evaluation criteria.}
Participants independently evaluated each generated scene along three complementary dimensions: realism, aesthetics, and functionality. Each dimension was rated on an integer scale from 0 to 10, where a higher score indicated better performance. The following written definitions were provided to all participants.

\begin{itemize}
\item \textbf{Realism.}
Realism measures whether the generated scene resembles a plausible real-world indoor room. Participants were instructed to consider whether the overall room layout, furniture scale, object selection, spatial relationships, and placement of furniture and other objects were consistent with realistic indoor environments. Scenes containing implausible layouts, severe object intersections, inappropriate furniture scales, or placements that would be unlikely to occur in a real room were expected to receive lower scores.

\item \textbf{Aesthetics.}
Aesthetics measures whether the furniture and other objects are arranged in a visually pleasing, orderly, and stylistically coherent manner. Participants were instructed to consider whether the objects appeared deliberately arranged rather than randomly placed or cluttered, whether the spatial composition was balanced and organized, and whether the furniture, decorations, and other objects followed a consistent visual style.

\item \textbf{Functionality.}
Functionality measures whether the room layout and the placement of furniture and other objects sufficiently and completely support the activities specified in the functional prompt. Participants were instructed to consider not only whether the required furniture and objects were present, but also whether their spatial arrangement, orientation, accessibility, and surrounding free space made the requested activities practically executable.

\end{itemize}

A score of 0 indicated that the scene completely failed to satisfy the corresponding criterion, whereas a score of 10 indicated that the scene satisfied the criterion exceptionally well. Participants were instructed to assess the three dimensions independently. For example, a scene could be visually appealing but receive a low functionality score if its layout did not adequately support the activities described in the prompt. Similarly, a functional room could receive a lower aesthetics score if its furniture was disorganized or stylistically inconsistent.

\paragraph{Score aggregation.}
Because each participant evaluated an assigned subset of 20 prompt scenes rather than the complete evaluation set, we aggregated all valid ratings collected for each method and evaluation dimension. Let $\mathcal{R}_{m,d}$ denote the set of ratings collected for method $m$ under evaluation dimension $d$. The final score is computed as
\begin{equation}
S_{m,d}
=
\frac{1}{\left|\mathcal{R}_{m,d}\right|}
\sum_{(i,j)\in\mathcal{R}_{m,d}}
s_{m,d}^{(i,j)}.
\label{eq:user_study_score}
\end{equation}
Here, $\mathcal{R}_{m,d}$ denotes the set of valid ratings collected for
method $m$ under evaluation dimension $d$, and
$s_{m,d}^{(i,j)}$ denotes the score assigned by participant $j$ to scene $i$
generated by method $m$. Ratings were grouped according to the anonymous
method identifiers and averaged across the assigned scenes and participants.
The resulting mean scores are reported as \textit{Real.}, \textit{Aesth.},
and \textit{Func.} in Table 2 of the main paper.

\section{Additional Experimental Results}
\label{sec:supp_additional_results}

\subsection{Qualitative analysis of construction strategies.}

\begin{figure*}[hb]
    \centering
    \includegraphics[width=\textwidth]{Figures/ablationV3.pdf}
    \caption{
    Qualitative comparison of agentic construction strategies under
the same functional prompt. One-shot generation produces a sparse
scene with noticeable collisions and out-of-bound objects; staged
construction adds major functional objects and improves geometric
validity; recursive construction further yields more complete
functional zones, richer local object compositions, and better
spatial organization.
    }
    \label{fig:dsl_ablation_supp}
\end{figure*}

Figure~\ref{fig:dsl_ablation_supp} compares one-shot, staged,
and recursive construction under the same functional prompt.
The one-shot result is relatively sparse, containing only a
desk and a small number of lounge objects, and fails to fully
support book access, reading, and quiet relaxation. Because
all objects are planned in a single pass, it also exhibits clear
geometric issues, including furniture extending beyond the
room boundary and local object collisions. Staged construction
adds a bookshelf, reading furniture, and more support-surface
objects while reducing collisions and out-of-bound placements,
although the central area remains crowded and the activity
zones are not clearly separated. Recursive construction further
produces a more structured layout, with a dedicated study area
on the right, a reading and relaxation area in the center, and a
large bookshelf at the back for book access. It also adds local
details such as tabletop objects, side cabinets, and plants while
preserving more circulation space. These results show that
recursive coarse-to-fine construction improves not only
functional-zone and functional-object coverage, but also spatial
organization, boundary compliance, and collision avoidance.
The higher FOP of one-shot generation is mainly due to its
substantially smaller number of generated objects, which reduces
the chance of introducing irrelevant objects at the cost of much
lower functional coverage and poorer geometric validity.

\subsection{Qualitative Comparison on the SceneEval Benchmark}
\label{sec:supp_sceneeval_visual_comparison}

\begin{figure*}[!htbp]
    \centering
    \includegraphics[width=\textwidth]{Figures/scene_eval_qualitative_comparisonV2.pdf}
    \caption{
    Qualitative comparison on the SceneEval benchmark.
    Each row shows results generated from the same text prompt by different
    methods. Red boxes highlight representative object-generation,
    placement, or spatial-relation failures, while green boxes highlight
    the corresponding regions in our results. Compared with existing
    methods, our approach produces more plausible object geometry,
    scene-appropriate object configurations, and more consistent
    object--object and object--architecture relationships.
    }
    \label{fig:supp_sceneeval_comparison}
\end{figure*}

Figure~\ref{fig:supp_sceneeval_comparison} provides a qualitative
comparison on two representative SceneEval prompts. The highlighted
regions illustrate common failure modes involving object selection,
geometry, placement, and spatial relationships.

For the bedroom prompt, Scene Weaver places the tables in an
unstructured manner and produces two table-like objects without the
corresponding chair required by the prompt. Scene Smith generates a bed
whose headboard orientation is inconsistent with the surrounding
bedside furniture. Code as Room produces malformed chair geometry, with
the chair legs oriented incorrectly. Scene Code represents the requested
double bed as two visually separated single beds, resulting in an
implausible sleeping arrangement. HSM places the mini fridge near the
middle of the room without a clear reference to a wall or nearby
furniture. In contrast, our method generates a coherent bedroom layout
in which both major furniture and smaller objects have plausible scales,
positions, and functional relationships.

For the dining-room prompt, Scene Weaver places an additional shelf-like
structure above an existing cabinet, resulting in a redundant and
spatially implausible composition. Scene Smith generates both a pendant
light and an additional ceiling-mounted light in nearly the same region.
Code as Room produces too few chairs and places them inconsistently
around the dining table. In the Scene Code result, the central table
appears insufficiently supported and visually floats above the floor.
HSM generates several small objects on the floor rather than placing
them on the intended shelves or storage furniture. Our result instead
places the dining table, chairs, storage furniture, lighting, and smaller
objects in functionally appropriate locations, while maintaining
consistent support and relative spatial relationships.

Overall, the comparison shows that satisfying a prompt requires more
than generating the requested object categories. A successful scene must
also preserve plausible object geometry, support relations, orientation,
architectural grounding, and functional arrangements among both large
and small objects. Our recursive construction process provides explicit
context from previously instantiated objects, enabling later objects to
be placed with respect to valid support surfaces and existing spatial
relations.

\subsection{Runtime Distribution Across Scene-Generation Stages} 
\label{sec:supp_runtime_distribution}
\begin{table*}[t]
    \centering
    \small
    \setlength{\tabcolsep}{2.0pt}
    \renewcommand{\arraystretch}{1.05}

    \resizebox{\textwidth}{!}{
    \begin{tabular}{llll}
    \toprule
    \textbf{Method}
    & \textbf{Runtime category}
    & \textbf{Included operations}
    & \textbf{Share} \\
    \midrule

    \multirow{3}{*}{Code as Room}
    & \makecell[l]{Scene understanding and\\planning}
    & \makecell[l]{Scene recognition, object listing, region\\
      partitioning, relationship-graph construction, and\\
      spatial planning}
    & 30.76\% \\

    & \makecell[l]{Iterative scene generation\\and revision}
    & \makecell[l]{Initial scene generation, three rounds of layout\\
      adjustment, and modification of walls and\\
      secondary-object placeholders}
    & 39.74\% \\

    & \makecell[l]{Object and appearance\\construction}
    & \makecell[l]{Object descriptions and detailed geometry,\\
      small-object generation, PBR materials and\\
      textures, and lighting, camera, and rendering\\
      configuration}
    & 29.51\% \\

    \midrule

    \multirow{3}{*}{Ours}
    & \makecell[l]{Planning and decision\\making}
    & \makecell[l]{Qwen-based recursive planning and construction\\
      decisions}
    & 21.87\% \\

    & \makecell[l]{Object generation,\\indexing, and execution}
    & \makecell[l]{Object generation or retrieval, asset indexing,\\
      grounding, and deterministic execution}
    & 59.09\% \\

    & Other operations
    & \makecell[l]{Remaining pipeline operations not included in\\
      the two categories above}
    & 19.05\% \\

    \bottomrule
    \end{tabular}
    }

    \caption{
        Percentage breakdown of runtime across the major stages of
        Code as Room and our method.
    }
    \label{tab:supp_runtime_distribution}
\end{table*}
Table~\ref{tab:supp_runtime_distribution} presents the normalized runtime distribution of Code as Room and our method. We report percentages rather than absolute wall-clock times because absolute generation time can vary with scene complexity, hardware configuration, asset availability, caching, and model-service latency. The comparison is therefore intended to identify the dominant computational bottlenecks of the two pipelines, rather than to provide a direct measurement of absolute generation speed. For Code as Room, scene understanding and planning account for 30.76\% of the total runtime. Initial scene generation, three rounds of layout adjustment, and subsequent modification of walls and secondary-object placeholders together account for 39.74\%. Object generation, materials, textures, lighting, and rendering configuration account for the remaining 29.51\%. Thus, scene-level understanding, planning, generation, and revision collectively occupy 70.50\% of its runtime. This distribution shows that repeatedly constructing and modifying the global scene is a major computational component of the Code as Room pipeline. In particular, iterative scene generation and revision alone require nearly 40\% of the total runtime, as the system must revisit previous scene-level geometry and spatial relationships during multiple rounds of adjustment. In contrast, our method spends 21.87\% of its runtime on Qwen-based planning and decision making. Object generation, asset indexing, grounding, and deterministic execution constitute the dominant component, accounting for 59.09\%, while other operations account for 19.05\%. Our method does not introduce a separate stage for repeatedly regenerating and repairing the complete scene. Instead, each recursive construction decision is directly executed and committed to the partial scene before the next construction step. The comparison indicates that our method shifts the primary computational cost away from repeated global scene revision and toward the generation and grounding of actual scene content. Because the internal stages of the two methods are not strictly identical, the table should be interpreted as a comparison of method-specific runtime allocation rather than as a one-to-one stage-level efficiency comparison.

\subsection{Reference-Based Quantitative Evaluation}

\paragraph{Experimental Setup.}
We conduct reference-based evaluation on the standard 3D-FRONT benchmark\citep{9710195},
following the evaluation protocol of FreeScene\citep{Bai_2025_CVPR}.
The processed benchmark contains 4,041 bedroom scenes, 900 dining-room
scenes, and 813 living-room scenes. We adopt the same train/test split,
with 3,879/162 bedroom, 723/177 dining-room, and 621/192 living-room
scenes for training/testing, respectively, resulting in 531 held-out
test scenes in total. We do not use the 3D-FRONT training split for model training, and directly
evaluate on the held-out test scenes following the same benchmark protocol. We use the same rendering setup and evaluate
the generated scenes against the corresponding held-out reference scenes
using FID, KID, SCA, and iRecall. FID and KID measure distribution-level
similarity between generated and reference scenes, while SCA measures
their distinguishability, with values closer to 50\% indicating better
distribution alignment. iRecall evaluates how faithfully the generated
scenes satisfy the spatial relations specified by the input instructions.

\paragraph{Results.}
As shown in Table~\ref{tab:quantitative_comparison}, our method consistently achieves the best performance across all three room categories. In particular, it obtains lower FID and KID scores than prior methods, indicating that the generated scenes better match the reference distribution. Our method also improves iRecall across bedroom, dining room, and living room scenes, demonstrating stronger adherence to the specified scene requirements. The consistent gains across both distributional and instruction-following metrics suggest that our method produces scenes that are not only perceptually closer to the reference data, but also more faithful to the intended functional and spatial constraints.

\begin{table*}[t]
\centering
\caption{Quantitative comparison on bedroom, dining room, and living room scenes.
$\downarrow$/$\uparrow$ indicate lower/higher is better, respectively; for SCA, values closer to 50\% are better.
KID values are reported in units of $10^{-3}$.}
\label{tab:quantitative_comparison}
\resizebox{\textwidth}{!}{
\begin{tabular}{lcccccccccccc}
\toprule
\multirow{2}{*}{Method}
& \multicolumn{4}{c}{Bedroom}
& \multicolumn{4}{c}{Dining room}
& \multicolumn{4}{c}{Living room} \\
\cmidrule(lr){2-5}
\cmidrule(lr){6-9}
\cmidrule(lr){10-13}

& FID$\downarrow$
& KID$_{\times 10^{-3}}$$\downarrow$
& SCA$\%$& iRecall$\%$$\uparrow$
& FID$\downarrow$
& KID$_{\times 10^{-3}}$$\downarrow$
& SCA$\%$& iRecall$\%$$\uparrow$
& FID$\downarrow$
& KID$_{\times 10^{-3}}$$\downarrow$
& SCA$\%$& iRecall$\%$$\uparrow$ \\
\midrule

ATISS\citep{paschalidou2021atiss}
& 122.37 & 0.74 & 58.77 & 44.06
& 134.75 & 24.25 & 59.83 & 32.56
& 120.10 & 16.44 & 57.02 & 36.31 \\

DiffuScene\citep{tang2024diffuscene}
& 129.34 & 0.81 & 59.13 & 53.32
& 142.37 & 28.36 & 60.88 & 37.17
& 135.93 & 29.07 & 58.98 & 39.58 \\

InstructScene\citep{lin2024instructscene}
& 114.86 & 0.68 & 56.37 & 72.71
& 129.13 & 15.27 & 58.93 & 61.47
& 111.52 & 8.65 & 55.32 & 57.21 \\

FreeScene\citep{Bai_2025_CVPR}
& 108.00 & 0.21 & 53.16 & 81.40
& 125.32 & 10.68 & 55.59 & 75.01
& 108.22 & 3.87 & 54.05 & 71.81 \\

\midrule
\textbf{Ours}
& \textbf{102.23}& \textbf{0.16}& \textbf{52.61}& \textbf{85.76}& \textbf{113.58}& \textbf{7.94}& \textbf{53.34}& \textbf{80.83}& \textbf{100.17}& \textbf{1.42}& \textbf{52.91}& \textbf{77.45}\\

\bottomrule
\end{tabular}
}
\end{table*}

\section{Comparison with GPT-6 Sol, a Strong General-Purpose Baseline}
\begin{figure*}[!htbp]
    \centering
    \includegraphics[width=0.8\textwidth,
        height=0.4\textheight]{Figures/GPT6_Visual.pdf}
    \caption{
    Qualitative comparison between GPT-6 Sol and our method on Function-Room Generation.
    }
    \label{fig:GPT6_comparison}
\end{figure*}

\begin{table}
    \centering
    \begin{tabular}{ccccccccc}
    \toprule
       Method
        & FZC$\uparrow$
        & FOC$\uparrow$
        & FOP$\uparrow$
        & ACC$\uparrow$
        & SUP$\uparrow$
        & NAV$\uparrow$& OOB$\downarrow$& COL$\downarrow$ \\
       GPT-6 Sol  &  70.2&  75.4&  56.1&  52.5&  49.7&  97.5&  27.9& 36.6\\
       Ours
        & {82.5}& {86.3}& {65.7}& {69.1}& {68.8}& {100.0}& {11.2}& {14.5}\\
         \bottomrule
    \end{tabular}
    \caption{Quantitative comparison between GPT-6 Sol and our method on the
    Function-Room benchmark.}
    \label{tab:gpt6_comparison}
\end{table}
We further compare our method with GPT-6 Sol as a strong general-purpose
baseline. Given the same function-room prompts, GPT-6 Sol directly generates
executable Blender code to construct complete 3D scenes, without task-specific
training for Function-Room Generation.

As shown in Table~\ref{tab:gpt6_comparison} and
Fig.~\ref{fig:GPT6_comparison}, GPT-6 Sol already exhibits strong functional
and spatial priors, but our method achieves consistently better functional and
geometric performance. GPT-6 Sol still produces fine-grained structural errors,
such as misplaced chair legs and abnormal stool-wheel placement, indicating
that direct general-purpose generation remains insufficiently reliable for
Function-Room Generation. Nevertheless, its strong priors suggest that such
models could potentially serve as teachers for SFT or provide feedback signals
for RL within our framework.

\section{Limitations}

Despite supervised fine-tuning, multimodal language models may still hallucinate implausible objects, spatial relations, or functional arrangements, especially in uncommon or highly specialized room types. In addition, our framework does not explicitly guarantee physical feasibility, accessibility, or compliance with domain-specific safety constraints. Therefore, highly specialized or safety-critical rooms, such as medical facilities, may still require review and manual refinement by domain experts to ensure functional correctness, safety, and regulatory compliance. 

\end{document}

%% file: math_commands.tex
\usepackage{amsmath,amsfonts,bm}

\def\eqref#1{equation~\ref{#1}}

\def\1{\bm{1}}

\DeclareMathAlphabet{\mathsfit}{\encodingdefault}{\sfdefault}{m}{sl}
\SetMathAlphabet{\mathsfit}{bold}{\encodingdefault}{\sfdefault}{bx}{n}

